\documentclass[10pt,twocolumn,letterpaper]{article}
\usepackage{graphicx}
\usepackage{amsmath}
\usepackage{amssymb}
\usepackage{booktabs}
\usepackage{enumitem}
\usepackage{indentfirst}
\usepackage{multirow}
\usepackage{multicol}
\usepackage[accsupp]{axessibility}
\usepackage{float}
\usepackage{bbding}
\usepackage{cvpr}              

\usepackage[dvipsnames]{xcolor}

\definecolor{cvprblue}{rgb}{0.21,0.49,0.74}
\usepackage[pagebackref,breaklinks,colorlinks,citecolor=cvprblue]{hyperref}

\def\paperID{11145} 
\def\confName{CVPR}
\def\confYear{2024}
\def\Methodname{OESSL}

\title{Mitigating Object Dependencies: Improving Point Cloud Self-Supervised Learning through Object Exchange}

\author{
Yanhao Wu $^{1}$~
Tong Zhang$^{2}$\href{mailto:tong.zhang@epfl.ch}{\Envelope} \quad
Wei Ke$^{1}$ ~
Congpei Qiu$^{1}$ ~
Sabine Süsstrunk$^{2}$ ~
Mathieu Salzmann$^{2}$\\
$^1$ School of Software Engineering, Xi'an Jiaotong University, China \\
$^2$ School of Computer and Communication Sciences, EPFL Switzerland \\
}

\begin{document}

\twocolumn[{%
\renewcommand \twocolumn[2][]{#1}%
\maketitle
\vspace{-2em}

\includegraphics[width=\textwidth]{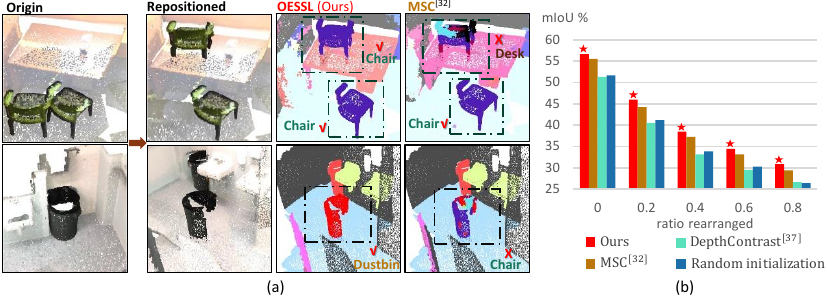}
\vspace{-2em}
\captionof{figure}{(a) Visualization of semantic segmentation for edited scenes. We relocate objects to places where they appear less frequently. Our pre-trained model segments the relocated object accurately, while the pre-trained model from MSC~\cite{MSC} labels the objects incorrectly. (b) Bar chart depicting the semantic segmentation performance on ScanNet~\cite{scannet} with varying ratios of rearranged objects. The X-axis indicates the ratios of rearranged objects for each scene, and the Y-axis shows the mean Intersection over Union (mIoU) scores. The models are pre-trained and fine-tuned on ScanNet with 10\% labels. We compare \Methodname~(Ours) with MSC~\cite{MSC}, DepthContrast~\cite{depthcontrast}, and training from scratch (weights are randomly initialized).\vspace{1em}} 
\label{fig:Teaser_Big}
}]

\begin{abstract}
\vspace{-1.5em}

In the realm of point cloud scene understanding, particularly in indoor scenes, objects are arranged following human habits, resulting in objects of certain semantics being closely positioned and displaying notable inter-object correlations. This can create a tendency for neural networks to exploit these strong dependencies, bypassing the individual object patterns. To address this challenge, we introduce a novel self-supervised learning (SSL) strategy. Our approach leverages both object patterns and contextual cues to produce robust features. It begins with the formulation of an object-exchanging strategy, where pairs of objects with comparable sizes are exchanged across different scenes, effectively disentangling the strong contextual dependencies. Subsequently, we introduce a context-aware feature learning strategy, which encodes object patterns without relying on their specific context by aggregating object features across various scenes. Our extensive experiments demonstrate the superiority of our method over existing SSL techniques, further showing its better robustness to environmental changes. Moreover, we showcase the applicability of our approach by transferring pre-trained models to diverse point cloud datasets. \footnote{Our code is available at \href{https://github.com/YanhaoWu/OESSL}{https://github.com/YanhaoWu/OESSL}. \\ \href{mailto:tong.zhang@epfl.ch}{\Envelope}~: Corresponding author}
\end{abstract}

\vspace{-0.5cm}

\section{Introduction}
\label{sec:intro}
\vspace{-0.2cm}
Understanding the semantic content of 3D point cloud data, particularly indoor scenes, is crucial in diverse fields, including applications such as indoor robotics~\cite{Mrke1, Robot_Segmentation2, Robot_Segmentation3, Robot_Segmentation4}. Recent advancements in deep learning~\cite{MinkUnet, octformer} have showcased remarkable results in this domain. While effective, these methods rely heavily on annotated training data and fail when faced with distribution shifts in the test data~\cite{Semantic-NeRF}. Consequently, the extraction of resilient object features from unlabeled data has become critical to advance the field.

Existing self-supervised learning (SSL) methods~\cite{STSSL, Mrke2, pointssl2, pointssl3, pointssl4} concentrate on feature aggregation by creating positive pairs from the same object in different augmented views of the scene. This maintains the relative relationships between objects unchanged, thus failing to account for the object dependencies.
Notably, in indoor point cloud scenes, object correlations are influenced by human habits, such as the association of tables with chairs, or toilets with sinks, resulting in strong inter-object entanglements. As demonstrated in Figure~\ref{fig:Teaser_Big}(a), the pre-trained models like~\cite{MSC} struggle to segment objects with unconventional correlations, such as chairs on desks or dustbins located away from walls. Although Mix3D~\cite{mix3d} has been proposed to augment the data by randomly combining two scenes, it does not reason at the level of objects.

Thus, the overlaps between objects introduced by this method can disrupt the coherent patterns formed by these objects. Without ground-truth labels, this disruption leads to less meaningful features, limiting the suitability of this approach in an SSL setting.

In this paper, our main focus is on developing an effective method for augmenting scene point clouds at the object level to mitigate the impact of human-induced biases in the context of self-supervised learning. Simultaneously, we aim to extract features that are more robust to varied inter-object correlations by better encoding both object patterns and contextual information.  
To this end, we introduce (i) an {\bf Object Exchange Strategy}: This approach involves exchanging the positions of objects of comparable size in different scenes. By doing so, we effectively break the strong correlations between objects while alleviating issues related to object overlap. (ii) A {\bf Context-Aware Object Feature Learning Strategy}: We first take the remaining objects, which share similar context in two randomly augmented views, as positive samples to encode the necessary contextual information and object patterns. To counter strong inter-object correlations, we minimize the feature distance between the exchanged objects in distinct contextual settings. Note that the contextual cues for a single object can vary significantly across scenes. Therefore, minimizing the feature distance between the exchanged objects enables the model to solely focus on out-of-context object patterns. These two components collectively provide a practical framework for learning robust features that encapsulate both object patterns and contextual information.

Furthermore, the exchanged objects 
may violate conventional human placement rules and appear incompatible with their environmental context. To effectively recognize such relocated objects, the model needs to comprehend both object patterns and context information. We therefore introduce an auxiliary task to enhance features related to both object and context. This task involves predicting which points belong to the objects that have been relocated. By engaging in this task, the model gains a more comprehensive understanding of both object patterns and contextual information.

Our contributions can be summarized as follows:
\begin{itemize}
\vspace{0.25em}
\item We introduce a novel point cloud \textbf{O}bject \textbf{E}xchange \textbf{S}elf-\textbf{S}upervised \textbf{L}earning framework, named~\Methodname,~for indoor point clouds that learn object-level feature representations by encapsulating both object patterns and contextual information. 
\vspace{0.5em}
\item We propose a novel object-exchanging strategy that breaks the strong correlations between objects without incurring object overlap. 
\vspace{0.5em}
\item We introduce an auxiliary task aimed at regularizing each object point feature to make it context-aware.

\end{itemize}

Our experiments on several datasets, including ScanNet~\cite{scannet}, S3DIS~\cite{S3DIS}, and Synthia4D~\cite{synthia4D}, demonstrate the effectiveness of our method, especially in terms of robustness to the contextual noise, as shown in ~\cref{fig:Teaser_Big} (b).

\vspace{-0.3cm}

\begin{figure*}[t]
  \centering
    \includegraphics[width=1\linewidth]{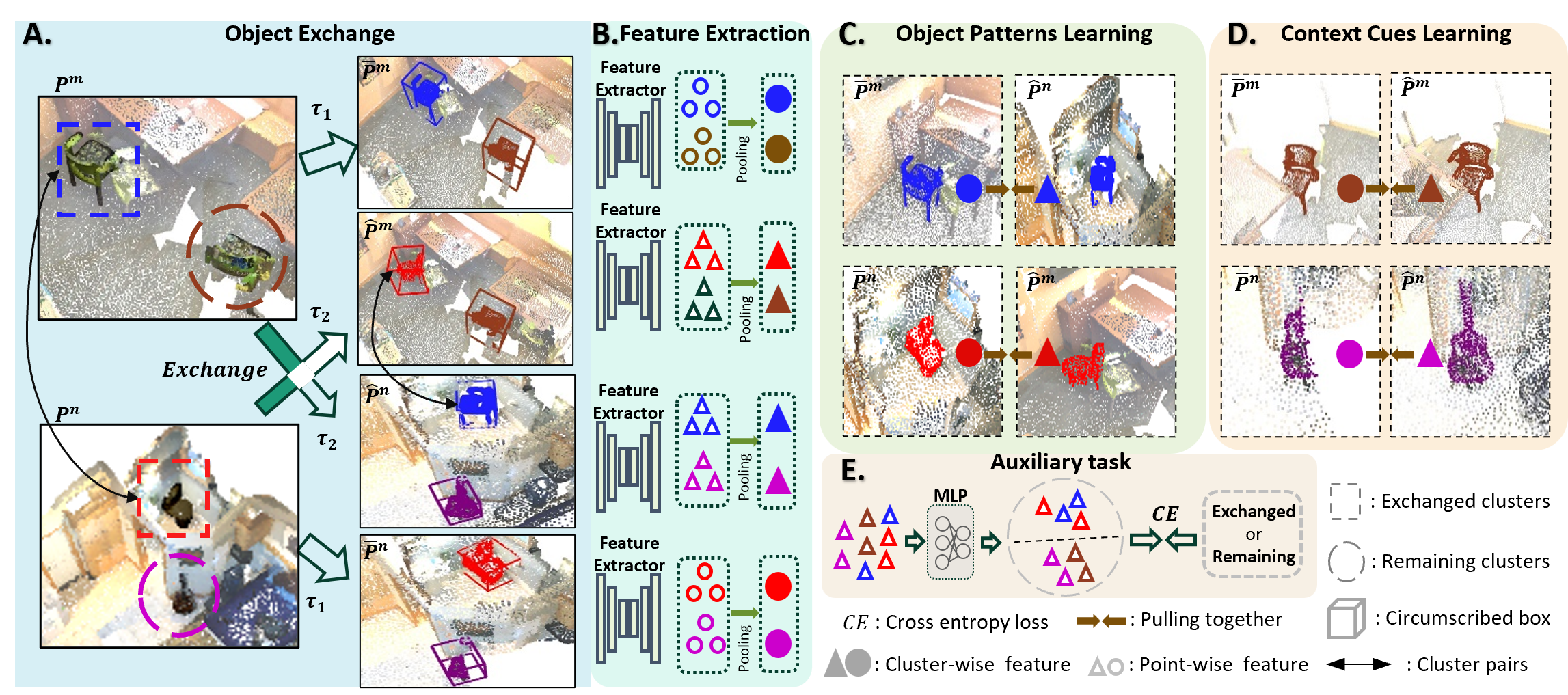}
    \vspace{-2em}
   \caption{{\bf Overview of our \Methodname}.~~\textbf{A}.~Given two randomly selected point clouds $P^{m}$ and $P^{n}$, we first perform clustering and generate minimum circumscribed boxes for every cluster. Clusters with similar circumscribed boxes are matched as cluster pairs. We exchange points of matched clusters and apply augmentation on $P^{m}$ and $P^{n}$ to generate novel views $\hat{P}^{m}$, $\hat{P}^{n}$, alongside two augmented views $\Bar{P}^{m}$ and $\Bar{P}^{n}$ without exchange.~\textbf{B.}~Every scene is passed through a feature extractor (Backbone) to obtain point-wise and cluster-wise features.~\textbf{C.}~We minimize the cluster feature distance obtained from the exchanged clusters in the different scenes (i.e.,$\Bar{P}^{m}$ and $\hat{P}^{n}$,$\Bar{P}^{n}$ and $\hat{P}^{m}$).~\textbf{D.}~We maximize the feature similarity between the remaining clusters in the augmented scenes (i.e.,$\Bar{P}^{m}$ and $\hat{P}^{m}$, $\Bar{P}^{n}$ and $\hat{P}^{n}$).~\textbf{E.}~The point-wise features are passed through a multilayer perceptron~(MLP) to classify which points belong to the relocated objects. The cross-entropy loss is used for classification. $\tau_{1}$ and $\tau_{2}$ are data augmentations, such as random flipping and random clipping.}
   \label{fig:overview_method}
    \vspace{-1em}
\end{figure*} 

\section{Related Work}
\label{sec:Related_Work}
\vspace{-0.2cm}

{\bf Training with context data augmentation.}
For image data, some researchers propose to add new instances to scenes to generate diverse training samples~\cite{Image_add_cues1, Image_add_cues2, Image_add_cues3, Image_add_cues4, Image_add_cues5}. Conversely, \cite{Image_decrease_cues1, Image_decrease_cues2, Image_decrease_cues3, Image_decrease_cues4, ICLR2024RCC} suggest removing contextual cues as data augmentation can also improve the model performance. However, the techniques designed for images cannot be directly applied to point clouds due to their distinct data nature.

In the 3D domain, 4dcontrast~\cite{PointSSL_4DContrast} augments scenes with moving synthetic objects and encourages feature similarity between corresponding objects. However, 4dcontrast needs synthetic datasets to obtain shapes, and moving a single object introduces limited contextual diversity. Nekrasov\textit{~et al.}~\cite{mix3d} propose a data augmentation named Mix3D which involves directly combining two point clouds and training models using augmented scenes in a supervised manner. The merged scene becomes chaotic with occlusions and overlaps, hindering the extraction of object-level features in the self-supervised learning (SSL) setting. Additionally, this exchange lacks meaningful object interactions and disrupts contextual information. By contrast, our object exchange strategy integrates objects from different scenes, greatly increasing the diversity of contextual cues while alleviating object overlap.




{\bf Self-supervised learning for 3D point clouds.}
Self-supervised learning for point clouds has developed rapidly in recent years~\cite{pointssl1, pointssl2, pointssl3, pointssl4, pointssl5}. 
In indoor scenes, recent research~\cite{pointcontrast, MSC, segcontrast} explores the nature of 3D point cloud data by aggregating features within the same point/object. For example, Pointcontrast~\cite{pointcontrast} and MSC~\cite{MSC} aggregate spatial features by maximizing the similarity between corresponding point features; DepthContrast~\cite{depthcontrast} and STRL~\cite{STRL} aggregate features in each region and pull features from different views together. 
Although effective, the correlation between indoor objects is strongly influenced by human bias, resulting in strong entanglements between objects. Therefore, aggregating features from indoor objects may lead to the model overfitting to inter-object correlations and ignoring object patterns.

By contrast, our method disrupts the correlations between objects to mitigate the model's dependence on contextual information. Additionally, we introduce a context-aware object feature learning strategy that leverages both object patterns and contextual information.

\vspace{-0.2cm}

\section{Method}
\label{sec:method}

\vspace{-0.2cm}
The overall framework of our method is depicted in Fig.~\ref{fig:overview_method} and contains two parts: Object exchange and context-aware object feature learning. We discuss these components in detail below.
\vspace{-0.2cm}

\subsection{Object exchange}
{\bf Unsupervised clustering.}~Let us be given a series of point clouds $P = \left\{P^{1},P^{2},...,P^{T}\right\}$ depicting $T$ scenes, where
 $P^k = (X^{k}, C^{k} ) = \left\{   (x^{k}_{1}, c^{k}_{1}),(x^{k}_{2},c^{k}_{2})...,(x^{k}_{N_k},c^{k}_{N_k}) \right\} $  represents the $k$-th point cloud with $N_k$ 3D points  $x^{k}_i \in {\mathbb{R}^{3}} $ and corresponding RGB colors $c^{k}_i \in {\mathbb{R}^{3}}$. 
For each 3D point set $X^{k}$, we compute normals for each point following~\cite{scannet}. This process yields a set of $N_{k}$ point noramls $O^{k}=\left\{O_{1}^{k}, O_{2}^{k}, ..., O_{N_k}^{k} \right\}$, $O_{i}^{k} \in {\mathbb{R}^{3}} $. Then, these points are taken as vertices to construct a graph whose weight matrix is defined as:
\vspace{-0.3cm}

\begin{equation}
    D = 2 - (D_{nor} + \alpha * D_{feat})\;,
    \label{eq:Clustering}
\end{equation}
where $D_{nor}$ represents the matrix of pairwise cosine similarity between the normals of two points, while $D_{feat}$ represents the matrix of pairwise cosine similarity based on point features. The parameter $\alpha \in [0, 1]$ serves as a weight, balancing the influence of the two matrices. We initialize $\alpha$ at 0 and iteratively update it during the feature learning process in~\cref{sec:Feature_Extraction}. Note that when the positions of two points, $i$ and $j$, are not spatially adjacent, $D_{ij}$ is set to a large number. Subsequently, we employ the GraphCut~\cite{graphcut} algorithm, a graph-based segmentation method, to cluster the points into $M_{k}$ clusters~\cite{scannet}.
The center of each cluster is determined as the average of all points belonging to that cluster.

{\bf Exchanging objects with comparable size.}~ To ensure meaningful object exchange without causing overlap with nearby objects, we adopt a systematic approach.  We first apply~\cite{convex_hull} to all the clusters to generate $M_{k}$ minimum circumscribed boxes, denoted as $B^{k}=\{B_{1}^{k}, B_{2}^{k}, ..., B_{M_k}^{k} \}$, where $B_{i}^{k}$ represents the length, width, and height of the $i$-th box in scene $k$. The pairwise box similarity is defined as the Euclidean distance between the vectors composed of length, width, and height, such that smaller distances correspond to higher similarity.
To enhance the diversity of exchanged objects, we employ a hybrid sampling strategy. For the $\beta M_{k}$ clusters in scene $k$, where $\beta$ is the preset exchange proportion of the clusters, we first select $ \dfrac{\beta}{2} M_{k} $ clusters using the farthest point sampling~(FPS) algorithm, ensuring a representative spatial distribution. The remaining clusters are then chosen via random sampling, introducing an element of randomness in the selection process.

Next, we introduce a similarity degree matrix, $V \in \mathbb{R}^{ \beta M_{k} \times M_{h}} $, where $V_{i, j}$ indicates the pairwise box similarity between cluster $i$ in scene $k$ and cluster $j$ in scene $h$. Following a greedy strategy, we match box pairs with the highest similarity in $V$. Subsequently, the points belonging to the corresponding matched clusters are exchanged between the two scenes. Leveraging $V$ helps to avoid object overlap, emphasizing the variability in contextual cues for a single object across different scenes. Further insights into the generation of robust features by exploiting such objects are discussed in~\cref{sec:Robust_feature}.

\subsection{Context-aware Object Feature Learning}

Having defined our object exchange strategy, we provide more detail on how to extract the features and establish our feature learning framework.
\vspace{-0.3cm}

\subsubsection{Feature extraction}\label{sec:Feature_Extraction}
Given an input point cloud $P^{n}$ and a randomly selected point cloud $P^{m}$ from the dataset, we apply our object exchange strategy and data augmentation to create two novel views $\hat{P}^{m}$ and $\hat{P}^{n}$, alongside two augmented views $\Bar{P}^{m}$ and $\Bar{P}^{n}$ without exchanging. To capture both point-wise and cluster-wise information, we leverage MinkUnet~\cite{MinkUnet} as our backbone encoder, denoted as $\phi$.

We initiate the feature extraction process by forwarding $\hat{P}^{m}$ through the backbone encoder, obtaining point-wise features $\hat{f}^{m}_{i}=\phi(\hat{P}^{m})$ for each 3D point. Organizing these features according to clusters results in a set of point-wise features, $\hat{F}^{m} = \{\hat{F}^{m}_{1}, \hat{F}^{m}_{2}, ..., \hat{F}^{m}_{M_{m}} \}$, where $\hat{F}^{m}_{i} \in \mathbb{R}^{N_{m,i} \times d}$, with $N_{m,i}$ representing the number of points in cluster $i$ from point cloud $\hat{P}^{m}$, and $d$ is the feature dimension of each $\hat{f}^m_i$. Additionally, we employ max-pooling on point features based on the clusters obtained using GraphCut, generating cluster-wise features $\hat{C}^{m} = \{\hat{c}^{m}_1, \hat{c}^{m}_{2}, \cdots, \hat{c}^{m}_{M_{m}} \} $, where $\hat{c}^m_i \in \mathbb{R}^{1\times d}$. The features in the other scenes can be obtained in the same way, as shown in~\cref{fig:overview_method}.

\vspace{-0.3cm}

\subsubsection{Feature aggregation}\label{sec:Robust_feature}
Aiming at a balanced concurrent ratio among objects of different semantics, we operationalize our approach through two central strategies for aligning cluster features: Object Patterns Learning and Contextual Cues Learning, both detailed below. Furthermore, we introduce an auxiliary task dedicated to enhancing the encoder's awareness of whether an object's feature distribution is in an unconventional location. This design aims to mitigate the challenges associated with cluster-level feature alignment by having regularization on point-level distribution.

{\bf Object patterns learning.} To encourage the model to learn object patterns, we minimize the feature distance between the clusters/points in the same cluster in different scenes. Note that the contextual cues for a single object can vary significantly between different scenes. Minimizing the feature distance between exchanged objects enables the model to solely focus on object patterns.

Let $M^{ex}_{m}$ denote the number of exchanged clusters in $\hat{P}^{n}$ that are originally located in ${P}^{m}$. We define a loss function
\vspace{-1cm}

\begin{equation} \label{eq:exchange_m}
\vspace{-0.3cm}
\begin{split}
& L_{op}^{m} = \frac{1}{M^{ex}_{m}} \times \sum_{i=1}^{M^{ex}_{m}} ( \left\| \frac{\hat{c}^n_{i}}{\| \hat{c}^n_{i} \|_2} - \frac{\bar{c}^m_{i}}{\| \bar{c}_{i}^{m} \|_2} \right\|_2^2\  \\ &~~~~~~~~~~~~~~~~ + \frac{1}{N_{m,i}} \times \sum_{j=1}^{N_{m,i}}
\left\| \frac{\hat{f}^n_{i,j}}{\| \hat{f}^n_{i,j} \|_2} - \frac{\bar{c}^m_i}{\| \bar{c}^m_i \|_2} \right\|_2^2\ ),
\end{split}
\end{equation}
where $\Bar{c}^m_{i}$ and $\hat{c}^n_{i}$ are the cluster-level feature vectors of the same exchanged clusters in $\Bar{P}^{m}$ and $\hat{P}^{n}$, and $\hat{f}_{i,j}^n$ represents the features of point $j$ belonging cluster $i$ in $\hat{P}^{n}$.
The loss function $L_{op}^{n}$ for the point cloud $P^{n}$ can be obtained in the same way. 
We then employ the symmetrized loss 
{\begin{equation} \label{eq:exchange}
\begin{split}
& L_{op} = L_{op}^{m} + L_{op}^{n}.
\end{split}
\end{equation}

{\bf Contextual cues learning.} 
To learn contextual cues, we minimize the feature distance between the remaining clusters, which share similar contexts in two randomly augmented views. To constrain the feature of each point, we also minimize the distance between the point and the corresponding cluster features~\cite{STSSL}. 

Let $M^{re}_{m}$ denote the number of remaining clusters that have not been exchanged in ${P}^{m}$. We write a loss
{\begin{equation} \label{eq:remain_m}
\begin{split}
& L_{context}^{m} = \frac{1}{M^{re}_{m}} \times \sum_{i=1}^{M^{re}_{m}} ( \left\| \frac{\hat{c}^m_{i}}{\| \hat{c}_{i}^{m} \|_2} - \frac{\bar{c}^m_{i}}{\| \bar{c}^m_{i} \|_2} \right\|_2^2\  \\ 
&~~~~~~~~~~~~~~~~ + \frac{1}{N_{m,i}} \times \sum_{j=1}^{N_{m,i}}
\left\| \frac{\hat{f}^m_{i,j}}{\| \hat{f}^m_{i,j} \|_2} - \frac{\bar{c}^m_i}{\| \bar{c}^m_i \|_2} \right\|_2^2\ ),
\end{split}
\end{equation}
where $\Bar{c}^m_{i}$, $\hat{c}^m_{i}$ are the cluster feature vectors of the same remaining cluster in $\Bar{P}^{m}$ and $\hat{P}^{m}$, and $\hat{f}_{i,j}^m$ represents the feature of point $j$ belonging cluster $i$ in $\hat{P}^{m}$.
The loss function $L_{context}^{n}$ for the point cloud $P^{n}$ can be obtained in the same way. 
We then define the symmetrized loss 
\vspace{-0.2cm}
{\begin{equation} \label{eq:remain}
\begin{split}
& L_{context} = L_{context}^{m} + L_{context}^{n}.
\end{split}
\end{equation}
\vspace{-0.4cm}

{\bf Auxiliary task.}
The auxiliary task aims to enable the model to gain a more comprehensive understanding of both object patterns and contextual information.
For the point cloud $\hat{P}^{m}$, we define a vector $\hat{Y}^{m} = \left\{ \hat{y}^{m}_{1}, \hat{y}^{m}_{2},..., \hat{y}^{m}_{\hat{N}_{m}} \right\}$, where $\hat{y}^{m}_{i} \in \left[0, 1 \right]$ represents whether point $i$ belongs to an exchanged
cluster and $\hat{N}_{m}$ represents the number of points in $\hat{P}^{m}$. We forward the point features $\hat{F}^{m}$ to a multilayer perceptrons~(MLP) to obtain point-wise prediction $\hat{Z}^{m} = \left\{ \hat{z}^{m}_{1}, \hat{z}^{m}_{2},..., \hat{z}^{m}_{\hat{N}_{m}} \right\}$, where $\hat{z}^{m}_{i} \in \left\{ 0, 1 \right\}$. For the point cloud $\hat{P}^{n}$, we obtain $\hat{Y}^{n}$ and $\hat{Z}^{n}$ in a same way.
We then define a loss $L_{aux}$ encoding the standard cross entropy loss between $\hat{Y}^{m}$ and $\hat{Z}^{m}$, and $\hat{Y}^{n}$ and $\hat{Z}^{n}$. 

Hence, our complete loss is written as 
\vspace{-0.2cm}
\begin{equation}\label{eq:total_loss}
L_{total} = L_{context} + \lambda L_{op} + \gamma L_{aux},
\end{equation}
\vspace{-0.5cm}

where $\lambda$ and $\gamma$ are weights balancing the three loss terms. We set $\lambda$ to 1 and $\gamma$ to 2 in our experiments.

\section{Experiments}
\label{sec:experiments}
In this section, we first introduce our experimental settings, including the datasets, object exchange details, and implementation details. Then, we evaluate our pre-trained models on downstream tasks and analyze our framework. 

\subsection{Experimental Settings}
\label{sec:setting}

{\bf Datasets}. ScanNet~\cite{scannet} consists of 3D reconstructions of real rooms and comprises 1513 indoor scenes. We follow the setting in~\cite{MinkUnet} and use a training and validation set, including 1201 and 312 scenes, respectively. The training set is used for pre-training and fine-tuning. Our framework utilizes scene-level point clouds for pre-training. The Standford Large-Scale 3D Indoor Space (S3DIS)~\cite{S3DIS} dataset contains 6 large-scale indoor areas~\cite{MinkUnet}. We use  area5 as validation data and the remaining areas as training data. Synthia4D is a large dataset that contains 3D scans of 6 sequences of driving scenes. Following~\cite{MinkUnet}, we split the Synthia4D dataset into train/val/test sets including 19888/815/1886 scenes. 

{\bf Object exchange details}. 
To obtain better segmentations for object exchange and feature extraction,  we update the point features with our learned features to create the affinity matrix. We set the initial relative weight $\alpha$ to 0 in Eq. (\ref{eq:Clustering}) and update the clusters twice during the training process: first at one third and then at two thirds, by setting $\alpha$ to 0.5.
We set the similarity threshold in GraphCut~\cite{graphcut} to 1.5 and merge the clusters with fewer than 300 points. In each scene, the clusters with any side length of the corresponding box exceeding 3 meters or less than 0.2 meters are not used for exchange. 


{\bf Implementation details}. We use MinkUnet~\cite{MinkUnet} as the backbone feature extractor and build our framework on the basis of BYOL~\cite{BYOL}. DepthContrast~\cite{depthcontrast}, MSC~\cite{MSC}, STRL~\cite{STRL}, and training from scratch are reproduced with the same backbone as ours to have fair comparisons. We pre-train the backbone on ScanNet for 200 epochs. The learning rate is initially set to 0.036 with a cosine annealing scheme with a minimum learning rate equal to $0.036 \times 10^{-4}$. We use SGD with a momentum of 0.9 and a weight decay of 0.0004 following STSSL~\cite{STSSL}. We use 8 $\times$ GTX3090 GPUs for pre-training and the batch size for each GPU is 12, which leads to a total batch size of 96. 

{\bf Evaluation metrics}. We use the mean intersection over union~(mIoU) and the overall point classification accuracy (Acc) to evaluate point cloud semantic segmentation, and average precision (mAP, AP@50\%, AP@25\%) for instance segmentation.

\vspace{-0.1cm}
\subsection{Scene Understanding}
To evaluate the pre-training methods, we employ different numbers of labels to fine-tune the models. In line with previous methods~\cite{depthcontrast, STSSL}, we partition ScanNet, S3DIS, and Synthia4D into distinct regimes, each corresponding to different percentages of labeled data. Specifically, we downsample the training data to 
levels of 10\%, 20\%, 50\%, and 100\% for ScanNet and S3DIS, and 0.1\%, 1\%, 10\%, and 100\% for Synthia4D. To mitigate randomness, we downsample three different regimes for every percentage, fine-tune the models separately using each regime, and report the average performance.
The number of training epochs for every label regime can be found in the supplementary.

{\bf Indoor scene understanding}. To evaluate the improvement of our~\Methodname~on indoor scene understanding, we fine-tune the pre-trained model on ScanNet. 

\begin{table}[h]
\tabcolsep=0.25cm
\vspace{-0.2cm}

\centering
\begin{tabular}{ccccc}
\hline
                          & 10\% & 20\%    & 50\%         & 100\% \\ \hline                   
From Scratch                & 48.99 & 57.58   & 61.70         &71.11  \\
DepthContrast~\cite{depthcontrast} & 50.30 & 57.08     & 61.47   & 70.92 \\
STRL \cite{STRL}                & 46.94 & 58.94  & 61.85        & 71.03 \\ 
MSC \cite{MSC}              & 53.85 & 60.47 & 63.98        & 71.00 \\ \hline
\textbf{\Methodname~(ours) } & {\bf 54.37}    & {\bf 61.27}   & {\bf 64.56}       & {\bf 71.28} \\ \hline
\end{tabular}
\vspace{-0.1cm}

\caption{Pre-training on {\bf ScanNet} and evaluating the fine-tuned models in different label regimes on \textbf{ScanNet} for semantic segmentation. We report the mIoU.}
\label{table:finetune_at_scannet_segmentation}
\vspace{-0.3cm}

\end{table}

\begin{table*}[h]
\centering
\setlength{\tabcolsep}{6mm}
\begin{tabular}{cccc}
\hline
 mAP / AP@50 / AP@25            & 10\% & 20\%           & 50\%         \\ \hline
From Scratch                & 12.63 / 25.90 / 43.33 & 23.63 / 41.42 / 60.73    & 30.91 / 51.25 / 68.38    \\
MSC \cite{MSC}              & 13.42 / 27.30 / 44.82 & 23.90 / 42.28 / {\bf 61.48}    & 29.16 / 51.18 / 68.71       \\ \hline
\textbf{\Methodname~(ours) } & {\bf 15.30 / 30.60 / 49.94}    & {\bf 24.67 / 43.28 }/ 60.86       & {\bf 31.73 / 52.06 / 69.80} \\ \hline
\end{tabular}
\vspace{-0.2cm}

\caption{Pre-training on {\bf ScanNet} and evaluating the fine-tuned models in different label regimes on \textbf{ScanNet} for instance segmentation~\cite{PointGroup}. We report the mAP, AP@50, AP@25.}
\label{table:finetune_at_scannet_instance}
\vspace{-0.5cm}

\end{table*}

In Table~\ref{table:finetune_at_scannet_segmentation}, we show the semantic segmentation results obtained by fine-tuning with different percentages of training data. Our method achieves better mIou for all label regimes than MSC. Specifically, our method outperforms training from scratch by 5.38\% at a level of 10\% and MSC~\cite{MSC} by 0.8\% at a level of 20\%. 
In Table~\ref{table:finetune_at_scannet_instance}, we report the instance segmentation results driven by \emph{PointGroup}~\cite{PointGroup}. When using 10\% of the labels for fine-tuning, our method improves performance by 4.7\% in AP@50\% compared to the network without pre-training. This evidences that our pre-training framework is also beneficial for discriminating instances.

\begin{table}[h]
\tabcolsep=0.25cm

\centering
\begin{tabular}{ccccc}
\hline
                          & 10\% & 20\%    & 50\%         & 100\% \\ \hline
From Scratch                & 40.48 & 45.94   & 53.25        & 66.16  \\
DepthContrast\cite{depthcontrast} & 46.57 & 47.67     & 53.85   & 63.42 \\
STRL \cite{STRL}                & 36.99 & 46.13  & 55.11        & 64.71 \\ 
MSC \cite{MSC}              & 44.85  & 50.12 & 57.16        & 65.40 \\ \hline
\textbf{\Methodname~(ours) } & {\bf 49.22}    & {\bf 52.67}   & {\bf 61.79}       & {\bf 66.90} \\ \hline
\end{tabular}
\vspace{-0.2cm}

\caption{Pre-training on {\bf ScanNet} and evaluating the fine-tuned models in different label regimes on \textbf{S3DIS} for semantic segmentation. We report the mIoU.}
\label{table:finetune_at_S3DIS_segmentation}
\vspace{-0.6cm}

\end{table}

{\bf Indoor scene transferability}. 
The contextual information significantly differs across datasets, making it difficult to transfer contextual features between different datasets, especially between indoor and outdoor scenes. By contrast, the object patterns, such as color and shape, are commonly shared between objects. Our method generates more transferable features by encoding object patterns without relying on their specific context.

To demonstrate the transferability of the features learned via our method, we pre-train models on ScanNet and fine-tune them for semantic segmentation on S3DIS~\cite{S3DIS}. As shown in Table~\ref{table:finetune_at_S3DIS_segmentation}, our pre-trained model performs better than the other methods. Specifically, our method outperforms MSC~\cite{MSC} by 4.37\% in mIoU with 10\% of the labels. These results strongly confirm the effectiveness of our approach at extracting object features that remain robust to changes in the environment.

{\bf Outdoor scene transferability}. 
We further fine-tune the models pre-trained on ScanNet for semantic segmentation using Synthia4D~\cite{synthia4D}, a self-driving dataset with different contexts than indoor scenes. In Table~\ref{table:finetune_at_S4D_segmentation_TEST} and Table~\ref{table:finetune_at_S4D_segmentation_VAL}, we report the mIoU obtained by fine-tuning the models using Synthia4D. Our method outperforms the other methods consistently across all label regimes. Specifically, our \Methodname~outperforms MSC~\cite{MSC} by 2.33\% with 1\% of the labels in the test set. When utilizing only 0.1\% of the training data, all pre-trained models exhibit a substantial improvement compared to training from scratch. Notably, our method achieves the most significant improvement, resulting in an mIoU of 49.32\% when evaluated on the validation set. 
The improvements on S3DIS~\cite{S3DIS} and Synthia4D~\cite{synthia4D} show that 
the features learned by our method generalize better than those learned by other methods.

\begin{table}[h]
\tabcolsep=0.25cm
\vspace{-0.3cm}

\centering
\begin{tabular}{ccccc}
\hline
     & 0.1\% & 1\%    & 10\%         & 100\% \\ \hline                   
From Scratch                & 19.84 & 63.37   & 70.45         & 77.00  \\
DepthContrast~\cite{depthcontrast} & 46.11 & 66.25     & 70.49   & 75.21 \\
STRL~\cite{STRL}                & 39.64 & 65.59  & 69.45        & 77.33 \\ 
MSC~\cite{MSC}       & 47.11 & 66.42 & 73.15        & 77.25 \\ \hline
\textbf{\Methodname~(ours) } & {\bf 49.44}    & {\bf 68.75}   & {\bf 73.42}       & {\bf 77.48} \\ \hline
\end{tabular}
\vspace{-0.2cm}

\caption{Pre-training on {\bf ScanNet} and evaluating the fine-tuned models on {\bf Synthia4D} for semantic segmentation. The models are evaluated on the {\bf test set}. We report the mIoU.}
\label{table:finetune_at_S4D_segmentation_TEST}
\vspace{-0.2cm}

\end{table}
\vspace{-1em}

\begin{table}[h]
\tabcolsep=0.25cm

\centering
\begin{tabular}{ccccc}
\hline
                   & 0.1\% & 1\%    & 10\%         & 100\% \\ \hline                   
From Scratch                & 20.17 & 67.87   & 74.35         & 80.50  \\
DepthContrast~\cite{depthcontrast} & 46.23 & 71.66  & 74.00   & 78.56 \\
STRL \cite{STRL}    & 38.27 & 70.49  & 73.80  & 80.95 \\ 
MSC \cite{MSC}       & 46.42 & 71.58 & 75.53   & 81.05 \\ \hline
\textbf{\Methodname~(ours) } & {\bf 49.32}    & {\bf 74.17}   & {\bf 77.04}       & {\bf 81.31} \\ \hline
\end{tabular}
\vspace{-0.2cm}

\caption{Pre-training on {\bf ScanNet} and evaluating the fine-tuned models under different label regimes on {\bf Synthia4D} for semantic segmentation. The models are evaluated on the {\bf validation set}. }
\vspace{-0.5cm}
\label{table:finetune_at_S4D_segmentation_VAL}
\end{table}

\vspace{-0.2cm}


\subsection{Ablation study}
In this section, we dissect our \Methodname~and analyze each component. Unless explicitly stated otherwise, the model is pre-trained and fine-tuned on ScanNet.

\begin{figure*}[t]
  \centering
    \includegraphics[width=1\linewidth]{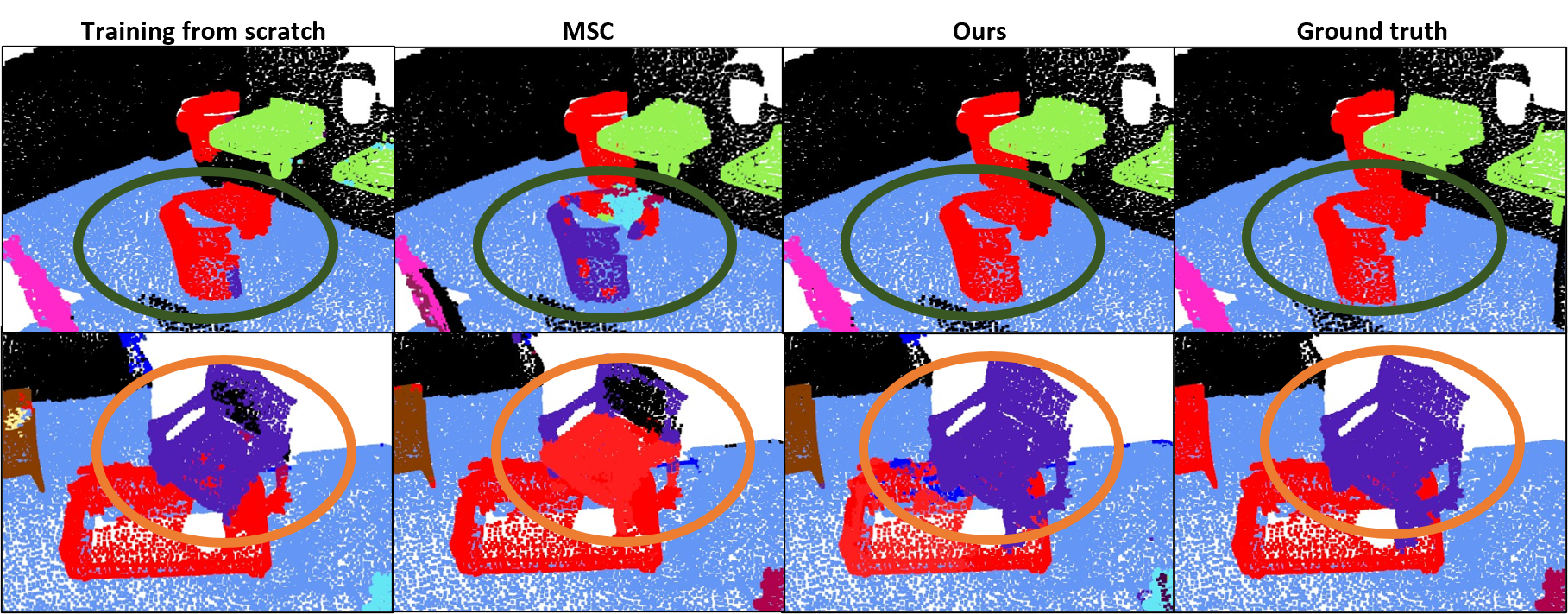}

           \vspace{-0.2cm}

   \caption{Segmentation results in scenes with objects relocated in unusual locations to eliminate contextual cues. We compare MSC~\cite{MSC}, \Methodname~(Ours), and training from scratch~(without pre-training). The model pre-trained with our method better distinguishes the relocated objects, as shown in the highlighted area (colored circles). }
   \label{fig:Contextual_Changes_Vis}
       \vspace{-0.4cm}

\end{figure*}

{\bf Breaking entanglements between objects.} Due to inherent human biases, strong correlations exist among indoor objects, indicating that certain classes of objects are highly likely to co-occur. This co-occurrence introduces the risk of the model overfitting to inter-object relations.

In Fig.~\ref{fig:AFF_MAP}, we illustrate the frequency of any two classes of objects appearing together. In the original training dataset (on the top of~\ref{fig:AFF_MAP}), certain classes exhibit a high frequency of appearing together. For instance, the shower curtain and door consistently appear simultaneously, and the co-occurrence frequency between the counter and cabinet is 0.9. However, by exchanging objects between scenes, our approach alleviates the high co-occurrence frequencies between objects, as shown in the bottom of Fig.\ref{fig:AFF_MAP}.

{\bf Performance under varied contexts.}
Our method avoids overemphasizing contextual cues and is therefore less affected by context changes compared to other SSL techniques. 
To validate this, we evaluate the model's performance in scenes with varied contexts. Specifically, we create a new dataset, ScanNet-C, by replacing a proportion $\delta$ of the objects in ScanNet with randomly selected objects from the entire dataset. We report the ratio of the model's performance on ScanNet-C to its performance on ScanNet. A higher ratio indicates a lower impact from contextual changes. In the experiment, we vary $\delta$ and repeat the experiment three times, reporting the average to reduce randomness. As shown in Table~\ref{table:performance_radio}, our pre-trained model consistently achieves higher mIoU values for all $\delta$ values, confirming that our method is indeed more robust to contextual changes than other methods.

In Fig.~\ref{fig:Contextual_Changes_Vis}, we visualize the semantic segmentation for scenes generated by relocating objects in a reasonable but unusual location. Specifically, a dustbin is placed far from the walls and a sofa is placed on the desk. For such objects with unreliable contextual cues, MSC~\cite{MSC} and the model without pre-training fail to segment the point clouds. By contrast, our \Methodname~accurately segments the objects, benefiting from object patterns learning. For additional visualizations and detailed information about ScanNet-C, please refer to the supplementary material.

\begin{table}[]
\setlength{\tabcolsep}{7pt}
\begin{tabular}{ccccc}
\hline
 Method~\textbackslash~$\delta$ & 0.2 & 0.4 & 0.6  & 0.8 \\ \hline
From Scratch        & 79.60                              & 65.50                              & 58.56                             & 51.01                              \\
DepthContrast~\cite{depthcontrast}  & 78.90                              & 64.55                             & 57.48                             & 51.88                              \\
MSC~\cite{MSC}            & 79.70                              & 67.05                              & 59.63                     & 52.92                              \\ \hline
\Methodname~(ours)           & \multicolumn{1}{l}{\textbf{80.99}} & \multicolumn{1}{l}{\textbf{67.75}} & \multicolumn{1}{l}{\textbf{60.67}} & \multicolumn{1}{l}{\textbf{54.43}}
\\ \hline
\end{tabular}
\vspace{-0.3cm}

\caption{{\bf Comparison of robustness to contextual changes.} We evaluate models on ScanNet-C with different proportions $\delta$ of replaced objects. We report the ratio(\%) of the model's performance on ScanNet-C to its performance on ScanNet.}
\vspace{-0.3cm}

\label{table:performance_radio}
\end{table}

\begin{figure}[t]
  \centering
    \includegraphics[width=1\linewidth]{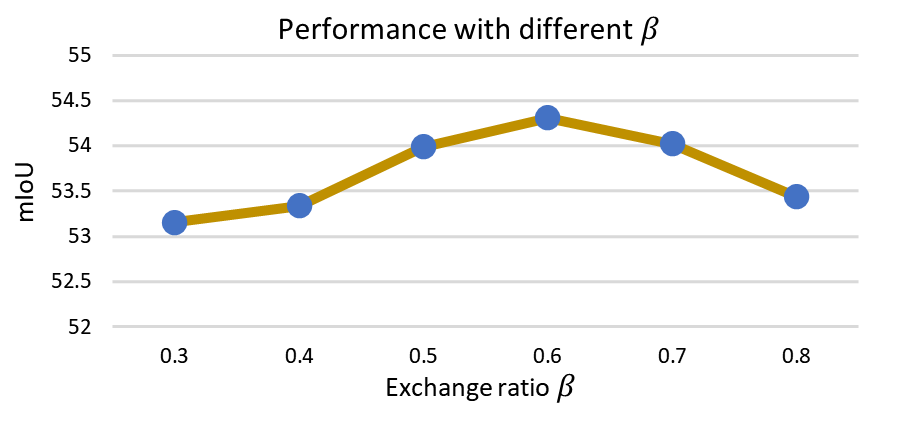}
    \vspace{-0.8cm}

   \caption{Comparison of mIoU on ScanNet, after fine-tuning the models pre-trained with different $\beta$.}
   \label{fig:Exchange_Radio_Lamuda}
       \vspace{-0.7cm}

\end{figure}

   

{\bf Effect of the exchanged object proportion.} In this study, we aim to clarify the impact of the exchange ratio on the learning process. The hyperparameter $\beta$ represents the proportion of exchanged clusters in the object exchange strategy. In our approach, when the number of available clusters in the scene exceeds 20, we set $\beta$ to 0.5; otherwise, we set it to 1. We keep $\beta$ fixed during pretraining to evaluate its impact on the model. The experiments are repeated three times to mitigate randomness. As depicted in Fig.~\ref{fig:Exchange_Radio_Lamuda}, the performance initially increases and then decreases as $\beta$ increases. We hypothesize that this is because a higher $\beta$ has the potential to increase the risk of object overlap, thereby completely disrupting existing contextual information. As shown in the bottom of Fig.~\ref{fig:Context_Aug_Method}, when
$\beta$ is set to 0.5, the desk is replaced by a bed, breaking the correlation between desk and sofa. However, a chair exchanges positions with the pillow on the sofa, disrupting the object patterns when $\beta$ equals 0.7. 
The best-performing model corresponds to setting $\beta$ to 0.6, which balances the number of exchanged objects and non-overlapping objects.

\begin{figure}[t]
  \centering
    \includegraphics[width=1\linewidth]{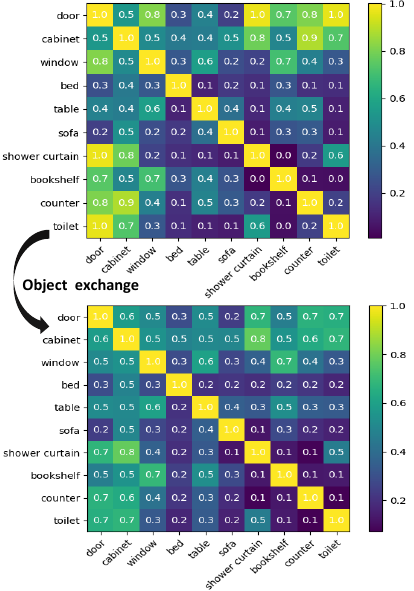}
       \vspace{-0.8cm}

   \caption{Affinity maps for the semantic classes in ScanNet~\cite{scannet}. Top: affinity map for the training set. Bottom: affinity map for the training set after object exchange.}
   
   \label{fig:AFF_MAP}
   \vspace{-0.1cm}

\end{figure}

   



\vspace{-0.3cm}

\begin{table}[h]
\setlength{\tabcolsep}{19pt}

\centering
\begin{tabular}{ccc}
\hline
                          & mIoU($\%$)       & Acc($\%$) \\ \hline
From Scratch                & 48.99       & 78.88  \\
MSC~\cite{MSC}     &53.85  & 80.49 \\ 
Baseline+Mix3D               & 52.62      & 80.19 \\ \hline
\textbf{\Methodname}              & \textbf{54.37}          & \textbf{81.15} \\ \hline
\end{tabular}
\vspace{-0.2cm}

\caption{Ablation study on the loss function with 10\% of the labels on ScanNet. We report mIoU/Acc.}
\label{table:ablation_on_outcontextual}
\vspace{-0.2cm}

\end{table}





\begin{figure}[t]
  \centering
    \includegraphics[width=1\linewidth]{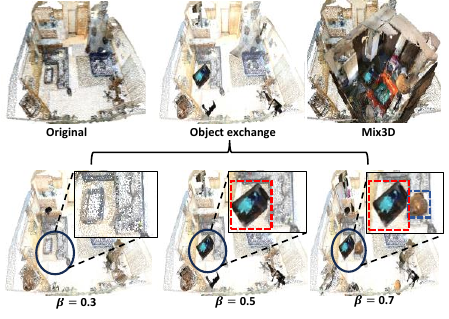}
   \caption{ Top: Visual comparison of scenes generated by Mix3D and our strategy. Bottom: Scenes generated by different $\beta$ using the object exchange strategy. When $\beta$ is set to 0.5, the desk is replaced by a bed (highlighted in the red box), but a chair is exchanged with the pillow (highlighted in the blue box) when $\beta$ increases to 0.7.
   For better visualization, we enhance the color contrast between objects from different scenes.}
   \label{fig:Context_Aug_Method}
   \vspace{-0.6cm}
\end{figure}




\vspace{-0.3cm}

{\bf Comparison with Mix3D.} Mix3D~\cite{mix3d} is an augmentation that directly combines two point clouds to generate novel scenes and is effective for supervised semantic segmentation training. Different from supervised training, self-supervised pre-training aims to generate structured embeddings. Specifically, the objects of the same class should be close in feature space and far from the objects from other classes. The overlap between objects incurred by Mix3D makes it difficult to distinguish the patterns between different object classes, resulting in an irregular feature space. Unlike Mix3D, our proposed object-exchanging strategy mitigates object overlaps, as shown in the top of Fig.~\ref{fig:Context_Aug_Method}.

\begin{table}[]
\centering
\setlength{\tabcolsep}{9pt}

\begin{tabular}{c|c|c|c|c}
\hline
Method                       & \multicolumn{1}{c|}{Context} & \multicolumn{1}{c|}{OP} & Aux & mIoU \\ \hline
Baseline                     & \checkmark                    &                                      &           & 53.12      \\
Baseline + $L_{OP}$              & \checkmark                    & \checkmark                           &           & 53.90     \\
\Methodname & \checkmark                    & \checkmark                           & \checkmark & {\bf 54.37} \\    \hline
\end{tabular}

\vspace{-0.2cm}

\caption{Ablation study on the {\bf loss functions} with 10\%~of the labels on ScanNet. {\bf Context}: Context cues learning, {\bf OP}: Object pattern feature learning, {\bf Aux}: Auxiliary task.}
\label{table:ab_on_module}
\vspace{-0.5cm}

\end{table}

To further highlight the effectiveness of our proposed object-exchanging strategy, we replace it with the Mix3D method and minimize feature distance between corresponding points/clusters in the newly generated scenes. This setting, referred to as Baseline+Mix3D in Table~\ref{table:ablation_on_outcontextual}, yields an mIoU of 52.62\%, lower than MSC and \Methodname. It implies that Mix3D is not suitable for self-supervised learning.

{\bf Loss functions.}~We ablate the three loss functions in Eq.~\ref{eq:total_loss} to validate their effectiveness. Initially, we set $\beta$ to 0, ensuring that only the remaining clusters contribute, and only the loss function of Eq.\ref{eq:remain} is applied. We refer to this configuration as the baseline. Subsequently, by adjusting $\beta$, we activate the loss function in Eq.\ref{eq:exchange}, specifically designed for object pattern learning. This setting is denoted as Baseline+$L_{OP}$. The results in Table~\ref{table:ab_on_module} show that Baseline+$L_{OP}$ outperforms the baseline, achieving an mIoU of 53.90\%. Our \Methodname~extends this by incorporating an auxiliary task, resulting in a remarkable mIoU of 54.37\%, demonstrating superior performance.

{\bf Different backbones.}
We conduct experiments using SPVCNN~\cite{SPVCNN} as the backbone. The results, presented in Table~\ref{table:backbone}, demonstrate the effectiveness of our method with SPVCNN~\cite{SPVCNN}.

\begin{table}[h]
\vspace{-0.3cm}
\centering
\setlength{\tabcolsep}{14pt}
\begin{tabular}{lcc}
\hline
 Method & mIoU(\%) & Acc(\%)   \\ \hline
From Scratch        & 45.59                      & 77.38                                                                                 \\
Baseline     & 47.38      & 78.68  \\ \hline
OESSL~(ours)        & {\textbf{49.02}} & {\textbf{79.25}} \\ \hline
\end{tabular}
\vspace{-0.3cm}
\caption{Ablation study on backbones. The models are pre-trained on ScanNet and tested with 10\% labels.}
\vspace{-0.5cm}
\label{table:backbone}
\end{table}

\vspace{-0.2cm}

\section{Conclusion}
\vspace{-0.3cm}

In this paper, we have introduced a SSL framework for point clouds, aiming to capture object features that are robust to noise and contextual variations. It starts by exchanging objects with comparable sizes between different scenes, breaking strong inter-object entanglements, and then learning both object patterns and contextual cues by leveraging exchanged and remaining objects. Altogether, our approach provides practical tools to learn robust context-aware representation features for indoor scenes. Our experiments evidence that our method outperforms the previous SSL methods for indoor point clouds.



{\small {\textbf{Acknowledgement.} This work was supported in part by the National Natural Science Foundation of China under Grant No. 62376209 and the Swiss National Science Foundation via the Sinergia grant CRSII5-180359.}
{
    \small
    \vspace{-0.3cm}

    \bibliographystyle{ieeenat_fullname}
    \bibliography{main}
}




%



\maketitlesupplementary


 \vspace{-0.6cm}

\setcounter{section}{0}

\section*{A. The relative weight of the auxiliary task loss.}
\label{sec:ablation}
\vspace{-0.2cm}

$\gamma$ is the relative weight of the auxiliary task loss in Eq.6 in the main paper. To study the impact of it, we gradually increase the relative weight $\gamma$. As shown in Fig.~\ref{fig:Exchange_Radio_gama}, with the increase of $\gamma$, the performance first increase and then decrease.

\begin{figure}[h]
  \centering
    \includegraphics[width=1\linewidth]{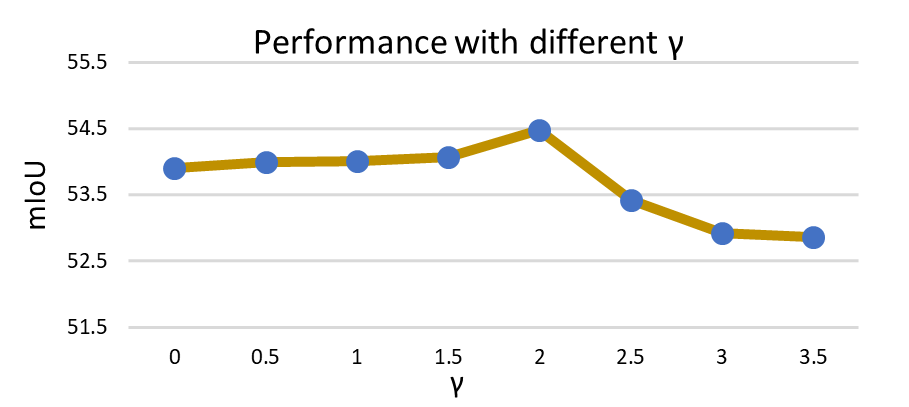}

   \caption{mIoU comparison under pre-training models with different $\gamma$. All the models are pre-trained and fine-tuned on ScanNet}
   \label{fig:Exchange_Radio_gama}
     \vspace{-0.5cm}
\end{figure}

\section*{B. Detailed ScanNet-C.}

\label{sec:ScanNet-C}
In Section 4.3 of the main paper, to evaluate the performance of models in changing contexts, we create a new dataset, ScanNet-C, by replacing a proportion $\delta$ of the objects in ScanNet. 

Specifically, for each point cloud $P^{m}$ with $N_{m}$ objects in ScanNet, we randomly select a point cloud $P^{n}$ with $N_{n}$ from the entire dataset. And then $\delta N_{m}$ objects in $P^{m}$ are replaced with objects sharing comparable size from $P^{n}$ using the object-exchanging strategy mentioned in the main paper. We replace objects in each point cloud in ScanNet and range $\delta$  from 0.1 to 0.9 in the experiments.  In Fig.~\ref{fig:scannet_c}, we visualize the scenes in ScanNet and the corresponding scenes in ScanNet-C. As shown in the figure, the inter-object correlations are changed, for example, a bed is replaced with a chair on the left of Fig.~\ref{fig:scannet_c}. In Table.~\ref{table:maintainig_performance}, we show each individual run on ScanNet-C semantic segmentation with varied proportions $\delta$. As the table shows, our \Methodname~outperforms all other methods under all $\delta$.

 \vspace{-0.2cm}

\section*{C. Detailed results and visualization.}
\label{sec:detailed results}

\begin{table}[h!]
\tabcolsep=0.34cm
\centering
\begin{tabular}{ccccc}
\hline
Label regime & 10\%  & 20\% & 50\% & 100\% \\ \hline
ScanNet~\cite{scannet}      & 250   & 250  & 100  & 75    \\
S3DIS~\cite{S3DIS}        & 400   & 300  & 200  & 200   \\ \hline
Label regime & 0.1\% & 1\%  & 10\% & 100\% \\ \hline
Synthia4D~\cite{synthia4D}    & 250   & 200  & 25   & 20  \\ \hline

\end{tabular}
\caption{Number of training epochs used for different label regimes on different datasets. }
\label{table:training_epoches}
\end{table}

The number of training epochs for every label regime can be found in Table~\ref{table:training_epoches}. For completeness, we report in Table.~\ref{table:scannet_s3d} and Table.~\ref{table:s4d} the mIoU of each of the three individual runs performed to obtain the main results in the paper. As the table shows, our method performs better than other methods consistently.

\begin{figure*}[h]
  \centering
    \includegraphics[width=1\linewidth]{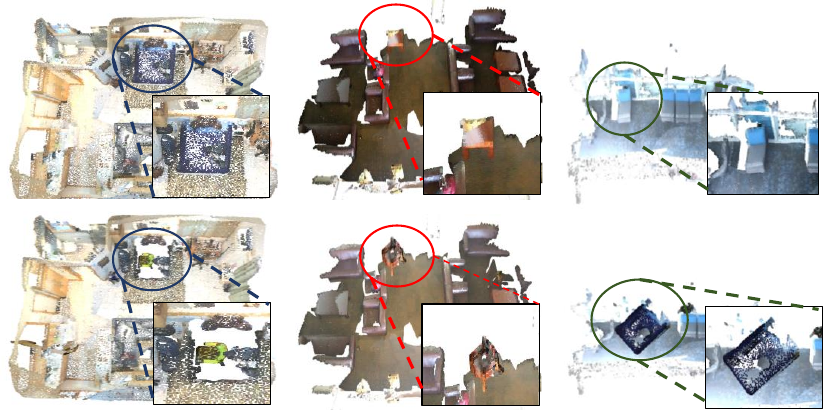}
   \caption{\textbf{Top}: Visualization of scenes in ScanNet. \textbf{Bottom}: Visualization of corresponding scenes in ScanNet-C}
   \label{fig:scannet_c}
\end{figure*}



\begin{table*}[h]
\begin{tabular}{c|c|cccccccccc}
\hline
Method        &         & 0     & 0.1   & 0.2   & 0.3   & 0.4   & 0.5   & 0.6   & 0.7   & 0.8   & 0.9   \\ \hline
From Scratch  &         & 51.73 & 46.51 & 40.66 & 37.82 & 34.09 & 30.79 & 30.43 & 27.60 & 26.38 & 26.29 \\
              & Runs    & 51.73 & 46.15 & 40.92 & 36.52 & 33.65 & 30.97 & 29.30 & 28.28 & 26.37 & 24.83 \\
              &         & 51.73 & 46.22 & 42.21 & 35.81 & 33.46 & 30.64 & 30.01 & 29.21 & 26.39 & 25.51 \\ \cline{2-12} 
              & Average & 51.73 & 46.29 & 41.26 & 36.72 & 33.73 & 30.80 & 29.91 & 28.36 & 26.38 & 25.55 \\ \hline
DepthContrast~\cite{depthcontrast} &         & 51.36 & 45.59 & 39.58 & 37.65 & 33.27 & 30.55 & 30.15 & 27.47 & 26.77 & 25.63 \\
              & Runs    & 51.36 & 45.67 & 40.15 & 36.59 & 33.18 & 30.28 & 28.80 & 27.95 & 26.46 & 25.14 \\
              &         & 51.36 & 45.15 & 41.84 & 34.90 & 33.02 & 30.71 & 29.61 & 28.76 & 26.71 & 25.48 \\ \cline{2-12} 
              & Average & 51.36 & 45.47 & 40.52 & 36.38 & 33.15 & 30.51 & 29.52 & 28.06 & 26.65 & 25.42 \\ \hline
MSC~\cite{MSC}           &         & 55.50  & 49.85 & 43.28 & 41.72 & 37.56 & 34.25 & 33.67 & 30.85 & 29.87 & 28.82 \\
              & Runs    & 55.50  & 49.68 & 43.95 & 40.74 & 36.86 & 33.60 & 32.44 & 31.19 & 29.20 & 27.98 \\
              &         & 55.50  & 49.49 & 45.48 & 39.07 & 37.22 & 34.10 & 33.17 & 32.40 & 29.05 & 28.70 \\ \cline{2-12} 
              & Average & 55.50  & 49.67 & 44.24 & 40.51 & 37.21 & 33.98 & 33.09 & 31.48 & 29.37 & 28.50 \\ \hline
OESSL(Ours)          &         & 56.72 & 51.54 & 44.98 & 42.95 & 38.30 & 35.82 & 35.46 & 32.10 & 31.32 & 29.86 \\
              & Runs    & 56.72 & 50.77 & 45.49 & 41.87 & 38.41 & 35.10 & 33.48 & 32.79 & 30.32 & 29.52 \\
              &         & 56.72 & 51.13 & 47.34 & 40.89 & 38.58 & 35.55 & 34.29 & 33.52 & 30.97 & 30.01 \\ \cline{2-12} 
              & Average & 56.72 & 51.15 & 45.94 & 41.90 & 38.43 & 35.49 & 34.41 & 32.80 & 30.87 & 29.80 \\ \hline
\end{tabular}
\caption{Detailed of individual runs on \textbf{ScanNet-C} semantic segmentation with different proportions~$\delta$ of replaced objects. We report mIoU\% for each of the individual runs averaged in the main paper.}
\label{table:maintainig_performance}
\end{table*}


\begin{table*}[h]
\centering
\begin{tabular}{cccccclccccc}
\hline
                     &                                    & \multicolumn{4}{c}{\textbf{ScanNet~\cite{scannet}}}                              &           &                      & \multicolumn{4}{c}{\textbf{S3DIS~\cite{S3DIS}}}                                \\
\multicolumn{1}{l}{} & \multicolumn{1}{l}{}               & \multicolumn{4}{c}{Validation}                                    &           & \multicolumn{1}{l}{} & \multicolumn{4}{c}{Area5}                                         \\ \hline
\%                   & \multicolumn{1}{c|}{Method}        & Split 1        & Split 2        & Split 3        & Average        &           &                      & Split 1        & Split 2        & Split 3        & Average        \\ \hline
10\%                 & \multicolumn{1}{c|}{From Scratch}  & 51.73          & 46.12          & 49.12          & 48.99          &           &                      & 35.32          & 41.86          & 44.27          & 40.48          \\
                     & \multicolumn{1}{c|}{DepthContrast~\cite{depthcontrast}} & 51.36          & 49.93          & 49.6           & 50.30          &           &                      & 45.10          & 47.84          & 46.76          & 46.57          \\
                     & \multicolumn{1}{c|}{STRL~\cite{STRL}}          & 50.29          & 48.00          & 42.52          & 46.94          &           &                      & 31.21          & 37.42          & 42.33          & 36.99          \\
                     & \multicolumn{1}{c|}{MSC~\cite{MSC}}           & 55.5           & 52.71          & 53.34          & 53.85          &           &                      & 43.61          & 48.46          & 42.48          & 44.85          \\
                     & \multicolumn{1}{c|}{OESSL(Ours)}   & \textbf{56.72} & \textbf{52.97} & \textbf{53.43} & \textbf{54.37} & \textbf{} & \textbf{}            & \textbf{46.71} & \textbf{49.88} & \textbf{51.07} & \textbf{49.22} \\ \hline
20\%                 & \multicolumn{1}{c|}{From Scratch}  & 55.22          & 57.78          & 59.73          & 57.58          &           &                      & 43.02          & 49.92          & 44.88          & 45.94          \\
                     & \multicolumn{1}{c|}{DepthContrast~\cite{depthcontrast}} & 55.81          & 57.59          & 57.83          & 57.08          &           &                      & 46.55          & 48.52          & 47.95          & 47.67          \\
                     & \multicolumn{1}{c|}{STRL~\cite{STRL}}          & 57.85          & 59.01          & 59.97          & 58.94          &           &                      & 44.48          & 49.6           & 44.44          & 46.13          \\
                     & \multicolumn{1}{c|}{MSC~\cite{MSC}}           & 59.67          & 59.85          & 61.88          & 60.47          &           &                      & 46.17          & 52.4           & 51.8           & 50.12          \\
                     & \multicolumn{1}{c|}{OESSL(Ours)}   & \textbf{60.33} & \textbf{60.58} & \textbf{62.91} & \textbf{61.27} & \textbf{} & \textbf{}            & \textbf{49.75} & \textbf{55.53} & \textbf{52.72} & \textbf{52.67} \\ \hline
50\%                 & \multicolumn{1}{c|}{From Scratch}  & 62.38          & 61.51          & 61.22          & 61.70          &           &                      & 51.27          & 53.51          & 54.97          & 53.25          \\
                     & \multicolumn{1}{c|}{DepthContrast~\cite{depthcontrast}} & 61.66          & 61.89          & 60.87          & 61.47          &           &                      & 52.86          & 53.55          & 55.14          & 53.85          \\
                     & \multicolumn{1}{c|}{STRL~\cite{STRL}}          & 61.78          & 62.38          & 61.38          & 61.85          &           &                      & 54.19          & 55.56          & 55.58          & 55.11          \\
                     & \multicolumn{1}{c|}{MSC~\cite{MSC}}           & \textbf{63.92} & 64.66          & 63.36          & 63.98          &           &                      & 56.56          & 56.48          & 58.43          & 57.16          \\
                     & \multicolumn{1}{c|}{OESSL(Ours)}   & 63.67          & \textbf{65.46} & \textbf{64.54} & \textbf{64.56} &           &                      & \textbf{60.98} & \textbf{61.95} & \textbf{62.43} & \textbf{61.79} \\ \hline
100\%                & \multicolumn{1}{c|}{From Scratch}  & 71.40          & 70.98          & 70.94          & 71.11          &           &                      & 65.54          & 66.18          & 66.75          & 66.16          \\
                     & \multicolumn{1}{c|}{DepthContrast~\cite{depthcontrast}} & 70.78          & 71.00          & 70.98          & 70.92          &           &                      & 63.68          & 61.18          & 65.41          & 63.42          \\
                     & \multicolumn{1}{c|}{STRL~\cite{STRL}}          & 70.38          & \textbf{71.56} & 71.15          & 71.03          &           &                      & 66.13          & 65.92          & 62.08          & 64.71          \\
                     & \multicolumn{1}{c|}{MSC~\cite{MSC}}           & \textbf{71.52} & 70.84          & 70.64          & 71.00          &           &                      & 65.83          & 63.55          & \textbf{66.83} & 65.40          \\
                     & \multicolumn{1}{c|}{OESSL(Ours)}   & 71.29          & 71.24          & \textbf{71.32} & \textbf{71.28} &           &                      & \textbf{67.55} & \textbf{67.49} & 65.65          & \textbf{66.90} \\ \hline
\end{tabular}
\caption{Details of individual runs on \textbf{ScanNet} and \textbf{S3DIS} semantic segmentation. Each run corresponds to fine-tuning using a different regime. We report mIoU\% for each of the individual runs averaged in the main paper}
\label{table:scannet_s3d}

\end{table*}

\begin{table*}[h]
\centering
\begin{tabular}{cccccclccccc}
\hline
                     &                                    & \multicolumn{4}{c}{\textbf{Synthia4D~\cite{synthia4D}}}                            &  &                      & \multicolumn{4}{c}{\textbf{Synthia4D~\cite{synthia4D}}}                            \\
\multicolumn{1}{l}{} & \multicolumn{1}{l}{}               & \multicolumn{4}{c}{Test}                                          &  & \multicolumn{1}{l}{} & \multicolumn{4}{c}{Validation}                                    \\ \hline
\%                   & \multicolumn{1}{c|}{Method}        & Split 1        & Split 2        & Split 3        & Average        &  &                      & Split 1        & Split 2        & Split 3        & Average        \\ \hline
0.1\%                & \multicolumn{1}{c|}{From Scratch}  & 16.81          & 21.92          & 20.79          & 19.84          &  &                      & 17.66          & 21.57          & 21.28          & 20.17          \\
                     & \multicolumn{1}{c|}{DepthContrast~\cite{depthcontrast}} & 48.87          & 44.69          & 44.78          & 46.11          &  &                      & 46.20          & 46.55          & 45.93          & 46.23          \\
                     & \multicolumn{1}{c|}{STRL~\cite{STRL}}          & 46.34          & 32.92          & 39.65          & 39.64          &  &                      & 43.67          & 41.37          & 29.77          & 38.27          \\
                     & \multicolumn{1}{c|}{MSC~\cite{MSC}}           & 49.51          & 45.58          & 46.24          & 47.11          &  &                      & 45.39          & 46.31          & 47.55          & 46.42          \\
                     & \multicolumn{1}{c|}{OESSL(Ours)}   & \textbf{52.56} & \textbf{48.13} & \textbf{49.62} & \textbf{49.44} &  &                      & \textbf{50.82} & \textbf{49.11} & \textbf{48.04} & \textbf{49.32} \\ \hline
1\%                  & \multicolumn{1}{c|}{From Scratch}  & 63.38          & 62.80          & 63.92          & 63.37          &  &                      & 67.74          & 67.77          & 67.92          & 67.81          \\
                     & \multicolumn{1}{c|}{DepthContrast~\cite{depthcontrast}} & 66.60          & 67.17          & 64.97          & 66.25          &  &                      & 71.14          & 71.57          & 72.27          & 71.66          \\
                     & \multicolumn{1}{c|}{STRL~\cite{STRL}}          & 67.67          & 64.88          & 64.23          & 65.59          &  &                      & 71.63          & 71.26          & 68.59          & 70.49          \\
                     & \multicolumn{1}{c|}{MSC~\cite{MSC}}           & 67.08          & 65.23          & 66.95          & 66.42          &  &                      & 72.93          & 71.83          & 69.98          & 71.58          \\
                     & \multicolumn{1}{c|}{OESSL(Ours)}   & \textbf{68.26} & \textbf{70.83} & \textbf{67.16} & \textbf{68.75} &  &                      & \textbf{73.88} & \textbf{74.66} & \textbf{73.98} & \textbf{74.17} \\ \hline
10\%                 & \multicolumn{1}{c|}{From Scratch}  & 71.84  & 68.75          & 70.76          & 70.45          &  &                      & 75.22          & 73.17          & 74.66          & 74.35          \\
                     & \multicolumn{1}{c|}{DepthContrast~\cite{depthcontrast}} & 69.31          & 70.82          & 71.33          & 70.49          &  &                      & 73.04          & 74.65          & 74.31          & 74.00          \\
                     & \multicolumn{1}{c|}{STRL~\cite{STRL}}          & 67.32          & 70.78          & 70.26          & 69.45          &  &                      & 75.54          & 72.92          & 72.95          & 73.80          \\
                     & \multicolumn{1}{c|}{MSC~\cite{MSC}}           & \textbf{72.64}          & 73.50          & 73.30          & 73.15          &  &                      & 75.52          & 74.96          & 76.10          & 75.53          \\
                     & \multicolumn{1}{c|}{OESSL(Ours)}   & 71.40          & \textbf{73.73} & \textbf{75.12} & \textbf{73.42} &  &                      & \textbf{76.60} & \textbf{77.16} & \textbf{77.37} & \textbf{77.04} \\ \hline
100\%                & \multicolumn{1}{c|}{From Scratch}  & \textbf{77.57} & 77.06          & 76.37          & 77.00          &  &                      & 80.71          & 80.74          & 80.06          & 80.50          \\
                     & \multicolumn{1}{c|}{DepthContrast~\cite{depthcontrast}} & 76.72          & 75.34          & 73.56          & 75.21          &  &                      & 76.88          & 79.44          & 79.36          & 78.56          \\
                     & \multicolumn{1}{c|}{STRL~\cite{STRL}}          & 77.34          & 76.53          & 78.11          & 77.33          &  &                      & 81.28          & \textbf{81.66} & 79.92          & 80.95          \\
                     & \multicolumn{1}{c|}{MSC~\cite{MSC}}           & 76.80          & 77.75          & 77.11          & 77.25          &  &                      & 80.84          & 80.78          & \textbf{81.52} & 81.05          \\
                     & \multicolumn{1}{c|}{OESSL(Ours)}   & 76.05          & \textbf{78.10}  & \textbf{78.29} & \textbf{77.48} &  &                      & \textbf{81.41} & 81.20          & 81.32          & \textbf{81.31} \\ \hline
\end{tabular}

\caption{Details of individual runs on \textbf{Synthia4D} semantic segmentation. Each run corresponds to fine-tuning using a different regime. We report mIoU\% for each of the individual runs averaged in the main paper.}
\label{table:s4d}

\end{table*}

\end{document}